\documentclass[letterpaper, 10 pt, conference]{ieeeconf}
\IEEEoverridecommandlockouts    
\usepackage{times}              
\usepackage{amsmath}            
\usepackage{amssymb}            
\usepackage{graphicx}
\usepackage{algorithm}
\usepackage[noend]{algpseudocode}
\usepackage{booktabs}
\usepackage{listings}
\usepackage{subfiles}
\usepackage{multicol}
\usepackage{bm}
\usepackage{placeins}
\usepackage{bbding}
\usepackage{multirow}
\usepackage{siunitx}
\let\labelindent\relax 
\usepackage{enumitem}
\usepackage{xspace}

\usepackage[abbreviations]{glossaries-extra}

\glssetcategoryattribute{abbreviation}{indexonlyfirst}{true}

\glssetcategoryattribute{abbreviation}{nohyperfirst}{true}

\newabbreviation{auroc}{AUROC}{Area Under the Receiver Operating Characteristic Curve}
\newabbreviation{accuracy}{Acc}{Accuracy}

\newabbreviation{cnn}{CNN}{Convolutional Neural Network}

\newabbreviation{fov}{FoV}{Field of View}
\newabbreviation{fpr}{FPR}{False Positive Ratio}

\newabbreviation{gnn}{GNN}{Graph Neural Network}
\newabbreviation{gcn}{GCN}{Graph Convolutional Network}
\newabbreviation{imu}{IMU}{Inertial Measurement Unit}
\newabbreviation{irl}{IRL}{Inverse Reinforcement Learning}

\newabbreviation{knn}{KNN}{K-Nearest Neighbors}

\newabbreviation{lagr}{LAGR}{Learning Applied to Ground Vehicles}
\newabbreviation{lidar}{LiDAR}{Light Detection and Ranging}

\newabbreviation{mlp}{MLP}{Multi-Layer Perceptron}
\newabbreviation{mpc}{MPC}{Model Predictive Controller}
\newabbreviation{mse}{MSE}{Mean Squared Error}

\newabbreviation{ood}{OOD}{out-of-distribution}

\newabbreviation{rbf}{RBF}{Radial Basis Function}
\newabbreviation{rmp}{RMP}{Riemannian Motion Policies}
\newabbreviation{ros}{ROS}{Robot Operating System}
\newabbreviation{ros1}{ROS~1}{Robot Operating System}
\newabbreviation{roc}{ROC}{Receiver Operating Characteristic}
\newabbreviation{rf}{RF}{Random Forest}

\newabbreviation{sdf}{SDF}{Signed Distance Field}
\newabbreviation{slam}{SLAM}{Simultaneous Localization and Mapping}
\newabbreviation{svm}{SVM}{Support Vector Machine}
\newabbreviation{svc}{SVC}{Support Vector Classifier}
\newabbreviation{wvn}{WVN}{Wild Visual Navigation}

\newabbreviation{vit}{ViT}{Vision Transformer}

\newabbreviation{bim}{BIM}{Building Information Modeling}

\makeatletter
\let\NAT@parse\undefined
\DeclareRobustCommand\onedot{\futurelet\@let@token\@onedot}
\def\@onedot{\ifx\@let@token.\else.\null\fi\xspace}
\makeatother

\newcommand{\etal}{et~al\onedot}

\makeatother
\usepackage{xcolor}
\usepackage{hyperref}
\hypersetup{
    colorlinks=true,
    citecolor={blue},
    linkbordercolor = {red},
    linkcolor={gray!50!black}, 
    citecolor={green!50!black},
    urlcolor={red},
}

\title{\LARGE \bf Diffusion Based Robust LiDAR Place Recognition}

\author{Benjamin Krummenacher, Jonas Frey$^{*}$, Turcan Tuna$^{*}$, Olga Vysotska, Marco Hutter
\thanks{ All authors are with Robotic Systems Lab~(RSL), ETH Zurich, Jonas Frey is also with Max Planck Institute for Intelligent Systems Tübingen, Germany.}
\thanks{$^{*}$ Indicates equal contribution.}%
\thanks{This project was supported by the Sony Research Grant 2023, the European Union’s Horizon 2020 research and innovation program under grant agreement No.852044, 101016970, and 101070405, EU Horizon 2021 program grant agreement No. 101070596, the NCCR digital fabrication, the ETH Zurich Research Grant No. 21, the SNSF project No.188596, the Max Planck ETH Center for Learning Systems, and Design++ fellowship program of ETH Zurich.}
}

\begin{document}
\thispagestyle{empty}
\pagestyle{empty}
\maketitle

\begin{abstract}
Mobile robots on construction sites require accurate pose estimation to perform autonomous surveying and inspection missions. Localization in construction sites is a particularly challenging problem due to the presence of repetitive features such as flat plastered walls and perceptual aliasing due to apartments with similar layouts inter and intra floors. 
In this paper, we focus on the global re-positioning of a robot with respect to an accurate scanned mesh of the building solely using LiDAR data. 
In our approach, a neural network is trained on synthetic LiDAR point clouds generated by simulating a LiDAR in an accurate real-life large-scale mesh.
We train a diffusion model with a PointNet++ backbone, which allows us to model multiple position candidates from a single LiDAR point cloud. 
The resulting model can successfully predict the global position of LiDAR in confined and complex sites despite the adverse effects of perceptual aliasing.
The learned distribution of potential global positions can provide multi-modal position distribution.
We evaluate our approach across five real-world datasets and show the place recognition accuracy of \SI{77}{\%} ($\pm$\SI{2}{m}) on average while outperforming baselines at a factor of $2$ in mean error.
\end{abstract}

\section{Introduction}
\label{sec:intro}
Autonomous mobile robots can become essential agents in construction sites to track the construction process and verify that the built structures match the building plans. 
The building plans for modern construction are typically provided by Building Information Modeling (BIM) models, which serve as a Digital Twin of the construction site. 
BIM models not only capture the geometry of the building but also detail the relationships between components and the timeline of the construction process.
For robots to relate their sensory information to the building plan and operate autonomously, the current position relative to a shared global reference frame is required. Some construction sites acquire regular LiDAR data collections and hence can provide a mesh of the construction to serve as a global map and reference frame. 

Often, robots do not know their initial position in indoor areas, known as the kidnapped robot problem~\cite{introduction_to_autonomous_mobile_robots}, given that satellite positioning is unavailable. Simultaneous localization and mapping (SLAM)~\cite{SLAM_essential_algo} techniques can be used; however, for re-localization, these methods require an initial guess.
A common solution is to facilitate visual markers with a known position~\cite{autonomous_construction_robots} or rely on a place recognition framework. However, visual marker installment is time-consuming, and existing place recognition methods do not generalize to the repetitive and confined environment of construction sites.
\begin{figure}[t]
  \centering
  \includegraphics[width=0.95\linewidth]{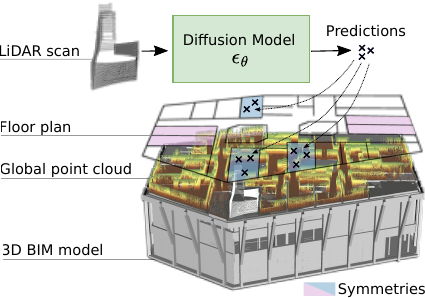}
  \caption{The proposed diffusion based LiDAR place recognition pipeline learns to predict potential positions in a given environment.}
  \vspace{-1em}
  \label{fig:motivation}
\end{figure}
\begin{figure*}[t]
  \centering
  \includegraphics[width=1.0\linewidth]{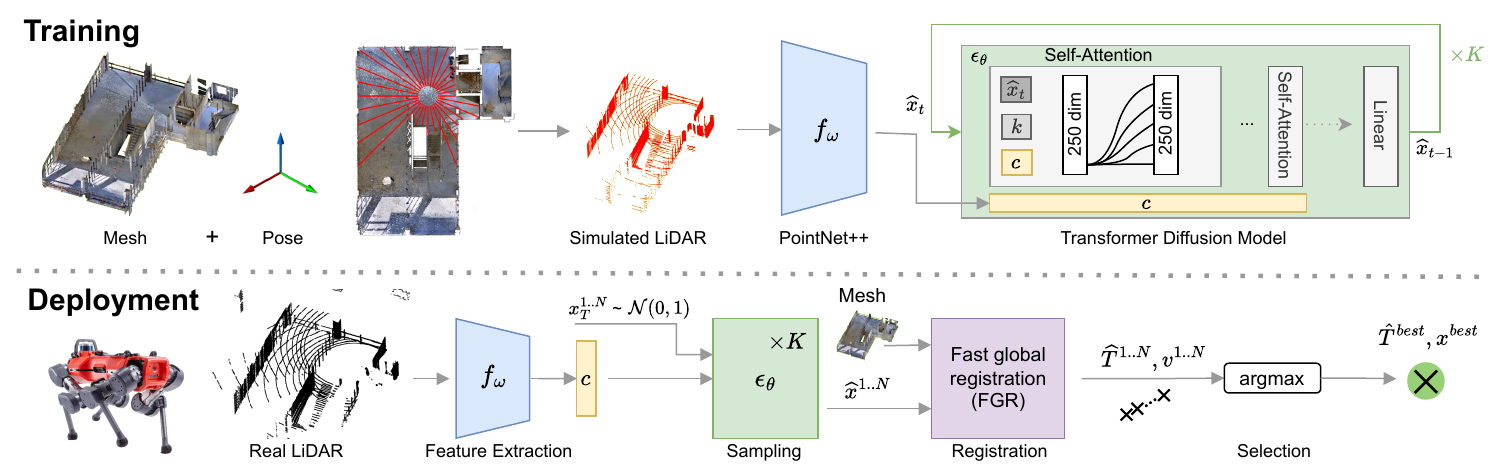}
  \caption{Overview of the proposed place recognition module. 
  During training, we sample random poses inside the mesh and ray cast the LiDAR pattern into the mesh.
  Features $\mathbf{c}$ are extracted from the resulting point clouds and provided as input to our diffusion model. During inference, a real LiDAR point cloud is passed into our model, which gradually denoises multiple random positions to $N$ candidate positions $\mathbf{\hat{x}}^{1..N}$. We use a diffusion model to generate multiple different candidate positions from one input point cloud. Fast global registration is used to select the best fitting candidate and refine the position estimate.}
  \label{fig:method_overview}
  \vspace{-1em}
\end{figure*}
To address this limitation, this work focuses on re-localizing a LiDAR point cloud in reference to a laser-scanned mesh of a building, i.e a mesh created from point clouds of a 3D laser scanner.
Firstly, we generate synthetic LiDAR scans by ray casting at randomly sampled positions in the mesh, which allows us to avoid the expensive real-world data collection step in construction sites.
Later, this training data is used to train a diffusion model that can predict the position of a LiDAR. We show that our framework, trained only on synthetic data, can be used to predict the position of a real LiDAR scan in the same environment despite the sim-to-real gap. 
Furthermore, we show that the diffusion model can predict potential positions in a multi-modal distribution, which is crucial in the presence of perceptual aliasing.
With our method, we can accurately recognize in which part of the construction site the robot is with an accuracy of up to $\pm$\SI{4}{\meter} in \SI{95}{\%} of the cases. In more complex scenarios with multiple similar rooms, the accuracy is around \SI{70}{\%}.

The key contributions of our work are:
\textit{(i)} A novel LiDAR-based place recognition module, based on a diffusion regression network, that is capable of handling complex construction site environments with perceptual aliasing, \textit{(ii)} synthetic LiDAR training dataset generation from reality capture-based mesh, \textit{(iii)} experimental evaluation across a wide variety of real-world construction datasets, and (iv) testing and thorough evaluation of the proposed module for downstream global registration applications as well as ablation study in localizability of the environment.
\section{Related Work}
\label{sec:related}
\textbf{LiDAR Place Recognition:}
Often, LiDAR place recognition methods use a database of global descriptors, handcrafted~\cite{keypoint_voting, scancontext} or learning-based descriptors~\cite{pointnetvlad, MinkLoc3D}, from LiDAR point clouds with known positions~\cite{lidarbasedplacerecognition_survey}. During deployment, the candidate descriptor is compared with the descriptors in the database to find the best match. Kim~\etal~\cite{scancontext} proposed the Scan Context, which splits the point cloud into bins and creates a 2D matrix using the maximal height from each bin. PointNetVLAD~\cite{pointnetvlad} is a learning-based LiDAR place recognition method. It combines the PointNet~\cite{pointnet} feature extraction with the NetVLAD~\cite{netvlad} global feature aggregation and is trained using a contrastive loss function. BEVPlace~\cite{bevplace} projects point clouds into bird's eye view images from which rotation-invariant features are extracted and aggregated into a global feature vector.
While the previous works consider the problem of matching LiDAR scans acquired by the same sensor (scan-to-scan), few works exist tackling the problem of scan-to-mesh. Qiao~\etal~\cite{lidar2bim} propose a method to match accumulated LiDAR submaps to a BIM model by segmenting walls from the submap. They then extract lines and corners and use a triangle descriptor to match these features to BIM submaps.

\textbf{Absolute Pose:}
Instead of finding the most similar LiDAR point cloud to a query point cloud, absolute pose regression (APR) directly finds the absolute pose of the point cloud. PointLoc~\cite{pointloc} is a LiDAR APR method using a PointNet++ point cloud encoder followed by a self-attention module. DiffLoc~\cite{diffloc} projects point clouds to range images and use a diffusion model to output the pose distribution. In contrast to our approach, they use tuples of LiDAR point clouds as input for their model. They also show that the variance across multiple predictions is highly correlated with positional error, which allows for the estimation of the uncertainty of a predicted pose.

\textbf{Diffusion Models:}
Diffusion models are generative models that build up on the ideas of energy-based models and allow the modeling of multi-modal probability distributions~\cite{denoisingdiffusionprobabilisticmodels}. 
They are trained by first applying noise to the data, typically Gaussian, within the forward process using a carefully crafted noise schedule in a multi-step procedure.
A neural network is then optimized to predict the applied noise, which can be subtracted from the noisy data, resulting in the reverse process consisting of individual denoising steps~\cite{denoisingdiffusionprobabilisticmodels}.
To sample from the learned distribution, during inference, the trained model is iteratively queried to remove noise starting from Gaussian noise~\cite{DenoisingDiffusionImplicitModels}. 
Diffusion models allow for learning joint distributions as well as conditional distributions, using classifier-free guidance~\cite{Jonathan2022}.
They achieved impressive results in a wide variety of fields~\cite{DBLP:journals/corr/abs-2112-10752, DBLP:journals/corr/abs-2103-01458,saharia2022photorealistictexttoimagediffusionmodels, yang2024diffusionmodelscomprehensivesurvey}. 
\section{Multi-Hypothesis LiDAR Place Recognition}
The overview of our place recognition pipeline is visualized in Fig.~\ref{fig:method_overview}. Our method can be separated into a training stage (pre-deployment) and a deployment stage. Our approach is able to generate an "almost infinite" amount of training data by exploiting the mesh of the environment. This allows us to generate any amount of synthetic data and train the model without requiring expensive robotic data collection and labeling beforehand. During training, we uniformly sample multiple 6-DoF pose $\mathbf{T} = (x, y, z, \phi, \theta, \psi)$ within the mesh of the building.
Then, we generate simulated point clouds by ray tracing \mbox{LiDAR} beams with the known sensor's scan pattern and specification. Once such a synthetic \mbox{LiDAR} scan is generated, we pass it to our diffusion position estimation model. Our model first extracts a feature vector using a PointNet++~\cite{pointnetPlusPlus} feature extractor. Then it employs the diffusion process to learn a distribution of $\hat{\mathbf{x}} = (x,y,z)$, conditioned on the extracted feature vector. This allows us to sample different positions, $\hat{\mathbf{x}}$. 

After training, the model can be deployed on the robot for inference within the deployment state. 
The robot perceives a \mbox{LiDAR} point cloud of the environment, which is then processed by a diffusion model, generating $N$ position predictions $\hat{\mathbf{x}}$ as initial guesses. 
This approach addresses perceptual aliasing by predicting a distribution across multiple modes corresponding to symmetrical rooms or floors.
We further use fast global registration (FGR)~\cite{FastGlobalRegistration} to select the best candidate position by registering the obtained point cloud at that position against point clouds simulated at the candidate positions. 

\subsection{Regression with Diffusion} 
Given a LiDAR point cloud $\mathbf{s_{\mathbf{x}}}$ generated at position $\mathbf{x}$ and a feature extractor $\bm{f}_{\omega}$, we first extract a feature vector from the point cloud $ \mathbf{c_{\mathbf{x}}} = \bm{f}_{\omega}(\mathbf{s_{\mathbf{x}}})$.
We want to learn the distribution $p(\mathbf{x} | \mathbf{c_{\mathbf{x}}})$ such that we can sample predicted positions from this distribution $\hat{p} \sim p(\mathbf{x} | \mathbf{c_{\mathbf{x}}})$, which is multi-modal in case of perceptual aliasing. 

To learn this distribution, we use a denoising diffusion probabilistic model~\cite{denoisingdiffusionprobabilisticmodels}, which introduces a diffusion process
$$q(\mathbf{x}_{1:T} \vert \mathbf{x}_{0}) := \prod_{t=1}^{T}q(\mathbf{x}_t \vert \mathbf{x}_{t-1})$$
$$q(\mathbf{x}_t \vert \mathbf{x}_{t-1}) = \mathcal{N}(\mathbf{x}_t; \sqrt{1 - \beta_t} \mathbf{x}_{t-1}, \beta_t\mathbf{I}) \quad$$ that consists of $T$ steps, where each step adds Gaussian noise to the ground truth position $\mathbf{x_0}$, resulting in the noisy positions $\mathbf{x}_1, \dots, \mathbf{x}_T$. 
The amount of noise added is defined in a variance schedule $\beta_1, \dots, \beta_T$~\cite{denoisingdiffusionprobabilisticmodels}.
We can sample any arbitrary step $\mathbf{x}_k$ in closed form
$$\mathbf{x}_k = \sqrt{\overline{\alpha}_t}\mathbf{x}_0 + \sqrt{1-\overline{\alpha}_t} \; \bm{\epsilon}, \: \bm{\epsilon} \sim \mathcal{N}(\mathbf{0}, \mathbf{I}).$$
with the notation $\alpha_t := 1- \beta_t$ and $\overline{\alpha_t} := \prod_{s=1}^{t} \alpha_s$~\cite{denoisingdiffusionprobabilisticmodels}.
During training, we train a model to reverse this diffusion process, hence removing noise from the noisy positions. 
We choose a random diffusion step $k \in \{1, \dots, T \}$, which we then use to sample a noisy position $\mathbf{x}_k$.
Then, we train a transformer model $\bm{\epsilon}_{\theta}$ to predict the noise $\hat{\bm{\epsilon}}$ added in the last step, given the time step embedding of $k$ and the global feature vector: $\hat{\bm{\epsilon}} \leftarrow \bm{\epsilon}_{\theta}(\mathbf{x}_k, \mathbf{c_{\mathbf{x}}}, k)$.

We use the mean absolute error between $\hat{\bm{\epsilon}}$ and the correct noise $\bm{\epsilon}$ as the loss function
$\mathcal{L}_{\text{1}}(\hat{\bm{\epsilon}}, \bm{\epsilon}) = \frac{1}{N} \sum_{i=1}^{N} \left| \hat{\bm{\epsilon}} - \bm{\epsilon} \right|
$ to optimize the weights $\omega$ and $\theta$.

During inference, the trained denoiser can start with an initial random position $\hat{\mathbf{x}}_T$ and iteratively compute less noisy positions~\cite{denoisingdiffusionprobabilisticmodels}
$$\hat{\mathbf{x}}_{t-1} = \frac{1}{\sqrt{\alpha_t}}(\hat{\mathbf{x}}_t - \frac{1-\alpha_t}{\sqrt{1-\overline{\alpha}_t}} \bm{\epsilon}_{\theta}(\mathbf{x}_{t},\mathbf{c_{\mathbf{x}}},k ))$$
until the predicted position $\hat{\mathbf{x}}_0$ is recovered. 
If we want multiple predictions from one point cloud, we can sample again with a different initial random position $\hat{\mathbf{x}}_T$. Multiple predictions can be inferred in parallel, given that we only have to infer $\bm{f}_{\omega}$ once to obtain $\mathbf{c_{\mathbf{x}}}$. 
\begin{figure*}[t]
  \centering
  \includegraphics[width=0.9\linewidth]{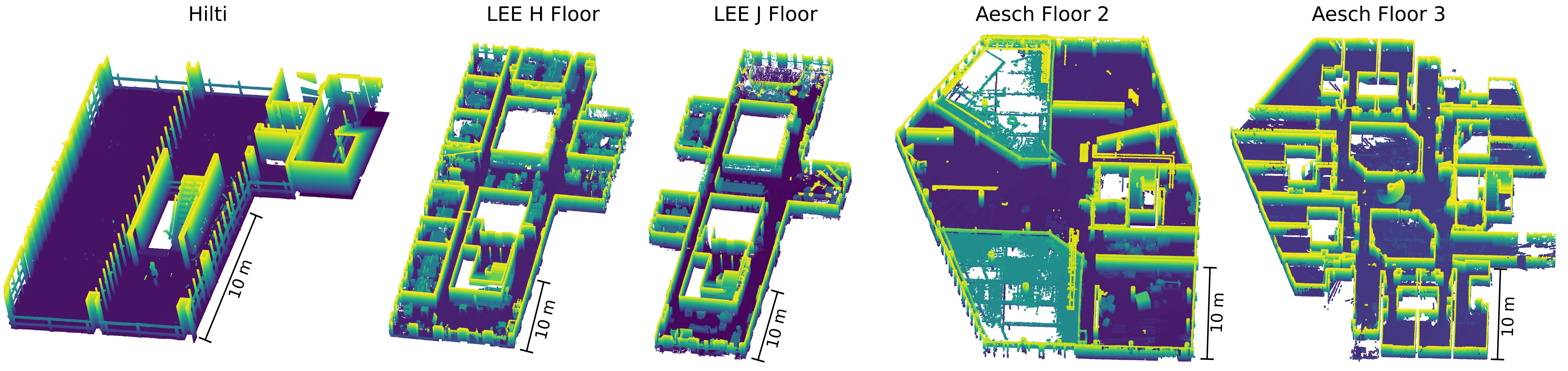}
  \caption{The datasets used in the evaluation are shown.}
  \label{fig:datasets}
  \vspace{-1em}
\end{figure*}

\subsection{Global Registration for Candidate Verification}
As an advantage of using a diffusion model, we can sample $N$ candidate positions from a single LiDAR point cloud.
This is particularly useful in the presence of symmetries in the environment, e.g., similar rooms on the same floor or even on multiple floors. However, to find a unique best fitting candidate position, we deploy fast global registration (FGR)~\cite{FastGlobalRegistration} to verify the potential candidate location for the fitness to the observed scene.
Fast global registration returns the transformation and a registration fitness value, which can be used to select the best registration result. To provide an even better initial guess, we apply the reported transformation to the best candidate position reported by our diffusion network.

\subsection{Implementation Details}
\textbf{Pre-Processing: }
Before passing the point clouds to the model, we remove all points near the LiDAR origin and all points in a 90-degree slice towards the back since, in our case, these points belong to the operator holding the platform. All points further from the origin than the maximal possible ray length in the dataset are removed since they mainly belong to neighboring buildings, which will differentiate the real from the simulated LiDAR point clouds. We center and normalize the position. To avoid numerical instabilities, the point cloud is also normalized by dividing all points by the maximal possible ray length, which ensures the relative sizes are preserved. Finally, we randomly subsample all point clouds to 4096 points to decrease the time needed for model training and inference.

\textbf{Model Architecture: }
Motivated by Wang~\etal~\cite{pointloc}, we use a PointNet++~\cite{pointnetPlusPlus} point cloud feature extractor. 
The hyperparameters of the feature extraction module were tuned using the Hilti dataset~\cite{Hilti2022} for validation.
The denoiser uses a sinusoidal embedding to embed the time step $t$ as a 256-dimensional vector. The denoiser consists of three parts: A linear layer with input dimension 256 + 256 + 3 and output dimension 256. The second part is a stack of 4 transformer encoder layers. Each transformer encoder layer takes 256 features as input, has four attention heads, and outputs 256 features. The third part is an MLP module with an input of 256 channels, a hidden layer of dimension 128, and an output dimension of 3.

\textbf{Training Details: } 
We sample rotations around axes perpendicular to the ground plane surface normal, uniformly between $-20$ and $20$ degrees.
We use $T = 100$ steps in the diffusion model and a linear variance schedule from $\beta_1 = 0.0001$ to  $\beta_T = 0.02$.
The model's weights are randomly initialized. 
We use a batch size of $16$ and the Adam optimizer with a learning rate of $0.0001$. We train the model for 900'000 steps, which takes 72 hours on an NVIDIA GeForce RTX 3060 GPU.
\begin{figure*}[t]
  \centering
  \includegraphics[width=0.95\linewidth]{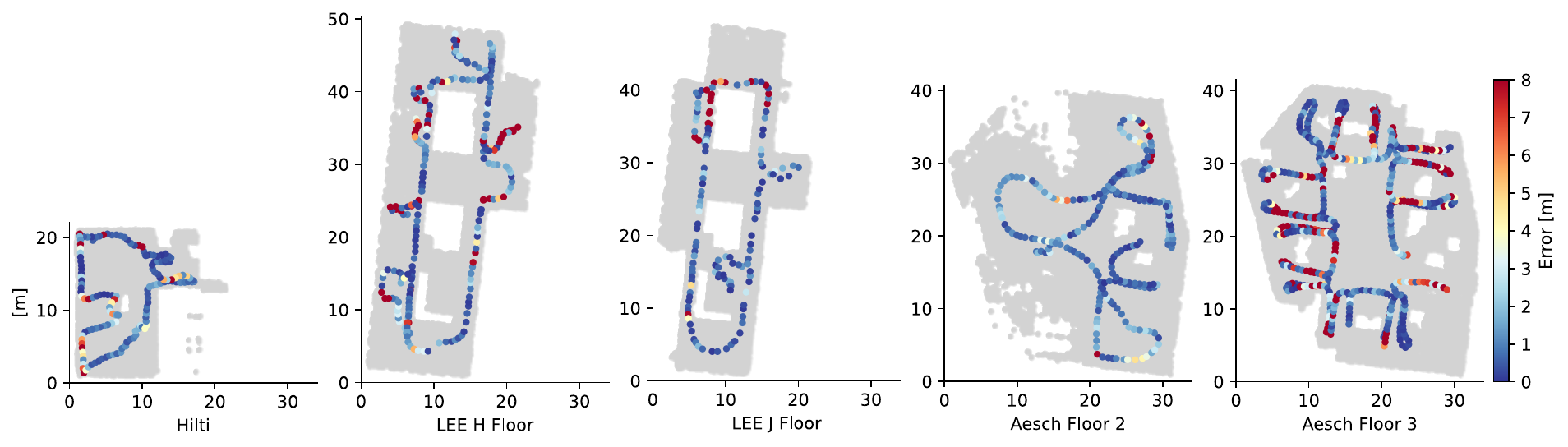}
  \caption{Trajectories of the different datasets are shown. The colors correspond to the error between the predicted and the ground truth positions. The grey area represents all the points sampled during training.}
  \label{fig:scatterplot}
  \vspace{-1em}
\end{figure*}
\section{Experiments and Evaluation}
\label{sec:exp}
\subsection{Datasets}
To illustrate the efficacy of our approach, we use three different datasets. Fig.~\ref{fig:datasets} visualizes all buildings and floors. 

\subsubsection{Hilti Dataset}
The Hilti dataset is from the Hilti 2022 challenge~\cite{Hilti2022}. We use trajectory exp06, a challenging sequence. A Hesai PandarXT-32 was used to collect the LiDAR, and a Z+F Imager 5016 3D stationary laser scanner was used to collect the global map.
\subsubsection{ETH LEE Building}
The LEE dataset is self-recorded, consisting of the two floors H and J in the ETH LEE building. These two floors consist of similar rooms and connecting corridors. Similarly, a Hesai PandarXT-32 LiDAR is used. To collect the global map, a Leica BLK2GO handheld laser scanner~\cite{blk2go_study} is used.
\subsubsection{Aesch Real Construction Site}
The Aesch dataset consists of the second and third floors of an active construction site. The second floor has a few big rooms with open office space, while the third floor consists of many small apartments, many of which have the same or a very similar layout. We collected the data with the LEE setup. This dataset is enabled by Design++ and ETH Zurich in collaboration with Halter AG.
\subsection{Baselines}
\begin{itemize}[leftmargin=0.3cm]
\item \textbf{Proposed}: We predict N=25 candidate positions and apply FGR to find the best position estimate.

\item \textbf{Grid Registration}: We generate simulated LiDAR scans in \SI{2}{\meter} uniform grid and use FGR to find the best position.
 
\item \textbf{Scan Context Grid}: We create the Scan Context database by simulated LiDAR scans in \SI{1}{\meter} uniform grid.

\item \textbf{Scan Context Trajectory}: We create the Scan Context database by simulated LiDAR scans along the ground truth trajectory.
\end{itemize}
We choose Grid Registration to measure the performance benefit of having targeted initial positions provided by our method compared to a naive grid search. The Scan Context Grid baseline is representative of a heuristic-based descriptor database method. We avoided using a learning-based place recognition method given the non-well-studied interaction of the contrastive learning objective and perceptual aliasing and the requirement for a large real-world LiDAR dataset.
With the Scan Context Trajectory, we provide privileged information to Scan Context, which allows us to verify the implementation. However, privileged information is not available in most real-world use cases. 
Moreover, we could not compare to \cite{lidar2bim} given that the code is not publicly available. 
\subsection{Single Floor Experiments}
In our first experiment, we evaluate the performance of our method in predicting position estimates in a single-floor scenario. 
In Fig.~\ref{fig:scatterplot}, we illustrate the prediction error between the predicted position of our method and the ground truth position for each sample along the trajectory. 

\begin{figure}[t]
  \centering
  \includegraphics[width=\linewidth]{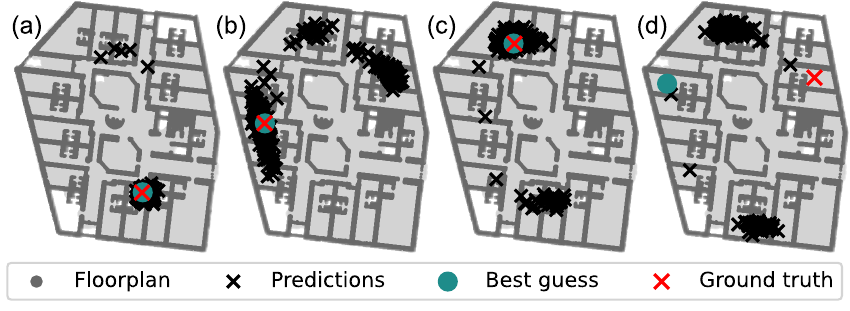}
  \caption{Distribution of candidate points. The best guess is found by performing fast global registration~\cite{FastGlobalRegistration}.}
  \label{fig:heatmap}
  \vspace{-2em}
\end{figure}

To demonstrate our method's ability to learn underlying multi-modal position distributions in cases of perceptual aliasing, we present four samples from the Aesch floor 3 in Fig.~\ref{fig:heatmap}. 
Our findings reveal that a scan recorded within symmetrical rooms (a, c) leads to a multi-modal distribution. 
A scan recorded within an entrance to a room with few robust features leads to a large variance across multiple entries in various rooms. 
The FGR can identify the correct location robustly.  
In (d), our method fails to identify the correct location and assigns the most probability mass to two wrong rooms. 
However, 3 samples are drawn from the correct symmetrical room pair, but in this case, FGR fails to identify the correct one. 

In Table~\ref{tab:single_floor}, we report the mean and median error, as well as the localization accuracy at the threshold of \SI{4}{\meter} and \SI{2}{\meter}.
A successful registration up to a threshold of \SI{2}{\meter} indicates that the correct room can be retrieved. 
Given the inherent symmetries within the building layout, as well as other perceptual aliasing that appear, e.g., when the LiDAR faces a wall, therefore we do not expect a \SI{100}{\%} accuracy. 

Our model predicts with an accuracy above \SI{60}{\%} at a threshold of \SI{2}{\meter} and an average mean error of \SI{2.84}{\meter} on all datasets. 
The model performs well in Hilti and Aesch Floor 2 datasets with \SI{4}{\meter} accuracies of \SI{95}{\%} and \SI{88}{\%} respectively, due to the absence of strong perceptual aliasing.
The Grid Registration baseline performs worse across all datasets, even though a candidate position is always within the \SI{2}{\meter} distance from the ground truth position. 
This error can be attributed to the scoring function used to assess the quality of registration of FGR, which selects the best candidate. 
Scan Context, when provided a database of descriptors from a \SI{1}{\meter} grid, does perform the worst even after our best attempts to tune the parameters. In our indoor scenario, most real LiDAR scans are mapped to a few single descriptors, which results in high errors. 

\begin{table}[b]
    \centering
    \renewcommand{\arraystretch}{1.1}
    \footnotesize
    \resizebox{\columnwidth}{!}{%
    \begin{tabular}{llccccc}
        \toprule
         Floor &  Approach  & mean [m] $\downarrow$& median [m]  $\downarrow$& acc@4m $\uparrow$& acc@2m $\uparrow$\\
         \midrule
         \multirow{4}{*}{Hilti} 
          & Ours & \textbf{1.53} &	\textbf{0.90} & \textbf{0.95} & \textbf{0.86} \\
          & Grid Registration&  4.89	&2.81	&0.61	&0.39 \\
          & Scan Context Grid&  7.54 &	6.72&	0.30&	0.24\\
          & Scan Context Traj. &  0.14	&9.75&	0.15&	0.04 \\
         \hline
         \multirow{4}{*}{LEE H} 
          & Ours & \textbf{3.69} & \textbf{0.59} & \textbf{0.77}	& \textbf{0.69} \\
          & Grid Registration & 7.66	&2.30&	0.58&0.48\\
          & Scan Context Grid &  16.23	&15.63&	0.11	&0.05 \\
          & Scan Context Traj. & 4.40 &	0.0&	0.74	&0.69\\
          
         \hline
         \multirow{4}{*}{LEE J} 
          & Ours & \textbf{2.56} & \textbf{0.35} & \textbf{0.87}	& \textbf{0.82}\\
          & Grid Registration & 6.96	&2.24&	0.62&	0.43\\
          & Scan Context Grid&  16.94	&15.31	&0.05	&0.05 \\
          & Scan Context Traj. &  3.75 &	0.0	&0.82 & 0.79\\
         \hline
         \multirow{4}{*}{Aesch 2}  
          & Ours & \textbf{1.22} &	\textbf{0.30} & \textbf{0.95} & \textbf{0.87 }\\
          & Grid Registration & 4.45	&1.97&	0.73&	0.50\\
          & Scan Context Grid&  16.95&	17.82	&0.11&	0.09 \\
          & Scan Context Traj. &  2.99 & 0.0	 & 0.79	& 0.73\\
         \hline
         \multirow{4}{*}{Aesch 3}  
          & Ours & \textbf{5.20} &	\textbf{0.95} & \textbf{0.69} & \textbf{0.61} \\
          & Grid Registration  & 9.36	&6.87 &	0.47&	0.37 \\
          & Scan Context Grid& 16.84 &	17.31	&0.04&	0.03 \\
          & Scan Context Traj. &  8.68 &5.86&0.45&0.41 \\
         \bottomrule
    \end{tabular}
    }
    \caption{single floor main result}
    \label{tab:single_floor}
\end{table}
\begin{table}[h]
    \centering
    \renewcommand{\arraystretch}{1.1}
    \resizebox{\columnwidth}{!}{%
    \begin{tabular}{llccccc}
        \toprule
         Floor &  Approach  &  mean [m] $\downarrow$& median [m]  $\downarrow$& acc@4m $\uparrow$& acc@2m $\uparrow$\\
         \midrule
         \multirow{2}{*}{LEE H} 
          & Ours & 4.98 & 1.87 & 0.74 & 0.54 \\
          & Grid Registration & 7.84 & 2.54 & 0.59 & 0.46 \\
         \hline
         \multirow{2}{*}{LEE J} 
          & Ours & 2.59 & 4.37 & 0.76 & 0.46\\
          & Grid Registration & 7.44 & 2.93 & 0.60 & 0.37 \\
         \hline
         \multirow{2}{*}{Aesch 2}  
          & Ours & 1.56 &0.84 & 0.95 & 0.84 \\
          & Grid Registration & 4.30 &1.97&0.74&0.50\\
         \hline
         \multirow{2}{*}{Aesch 3}  
          & Ours & 7.11 & 2.36 & 0.76 &	0.46 \\
          & Grid Registration  & 10.05 & 7.5 & 0.41& 0.35\\
         \bottomrule
    \end{tabular}
    }
    \caption{Results multi-floor}
    \label{tab:multifloor}
\end{table}

To validate our implementation and the capability of Scan Context to provide good estimates, we also rendered the simulated LiDAR scans exactly at the ground truth position.
Even in this simpler case, the Scan Context descriptor is not capable of coping with the sim-2-real gap between the simulated and real LiDAR scan. Even with privileged information, our method outperforms Scan Context Trajectory in accuracy. 
\begin{table}
    \centering
    \renewcommand{\arraystretch}{1.1}
    \resizebox{\columnwidth}{!}{%
    \begin{tabular}{llccccc}
        \toprule
         Floor &  Approach  &  mean [m] $\downarrow$& median [m]  $\downarrow$& acc@4m $\uparrow$& acc@2m $\uparrow$\\
         \midrule
         \multirow{3}{*}{Hilti} 
          & Ours (N=25) + FGR & \textbf{1.53} &	\textbf{0.90} & \textbf{0.95} & \textbf{0.86} \\
          & Non-diff. + FGR & 1.94 & 0.57 & 0.83 & 0.75 \\
          & Ours(N=1)  + FGR &  2.57	&1.1 &	0.83 &0.70 \\
          & Ours(N=25) + Oracle & 0.77 & 0.33 & 0.97 & 0.91 \\
          
         \hline
         \multirow{3}{*}{LEE H} 
          & Ours (N=25) + FGR& \textbf{3.69} & \textbf{0.59} & \textbf{0.77}	& \textbf{0.69} \\
          & Non-diff. + FGR & 4.27 & 1.54 & 0.63 & 0.54 \\
          & Ours(N=1)  + FGR&  6.04 & 2.86 & 0.61 &	0.33 \\
          & Ours(N=25) + Oracle & 0.96 & 0.54 & 0.97 & 0.91 \\
         \hline
         \multirow{3}{*}{LEE J} 
          & Ours (N=25) + FGR & \textbf{2.56} & \textbf{0.35} & \textbf{0.87}	& \textbf{0.82}\\
          & Ours(N=1)  + FGR & 4.69 & 1.58 & 0.79	& 0.63 \\
          & Non-diff. + FGR & 3.19 & 0.95 & 0.77 & 0.68 \\
        & Ours(N=25) + Oracle & 1.11 & 0.39 & 0.96 & 0.92\\
         \hline
         \multirow{3}{*}{Aesch 2}  
          & Ours (N=25) + FGR & \textbf{1.22} &	\textbf{0.30} & \textbf{0.95} & \textbf{0.87} \\
          & Non-diff. + FGR & 1.98 & 0.81 & 0.86 & 0.72 \\
          & Ours(N=1)  + FGR & 2.27 & 1.43 & 0.88 & 0.65 \\
        & Ours(N=25) + Oracle & 0.64 & 0.38 & 0.99 & 0.96 \\
         \hline
         \multirow{3}{*}{Aesch 3}  
          & Ours (N=25) + FGR & 5.20 &	\textbf{0.95} & \textbf{0.69} & \textbf{0.61} \\
          & Non-diff. + FGR & \textbf{5.09} & 1.37 & 0.61 & 0.53 \\
          & Ours(N=1)  + FGR &  6.76 & 2.72 & 0.61 & 0.38 \\
          & Ours(N=25) + Oracle & 1.51 & 0.62 & 0.92 & 0.86 \\
         \bottomrule
    \end{tabular}
    }
    \caption{Influence of multiple predictions }
    \vspace{-2em}
    \label{tab:ablation}
\end{table}
\subsection{Multifloor Experiments}
To show that our method can cope with multiple floors, we also trained our methods on two datasets, the merged LEE H and J Floors and the merged floors 2 and 3 of the Aesch dataset. The performance of these models is listed in Tab.~\ref{tab:multifloor}.
We can observe a slight degradation in performance when with the \SI{4}{\meter} threshold within the LEE H and J floor ($0.77\rightarrow0.74$ and $0.87\rightarrow0.76$).
A stronger decrease in performance is observed within the \SI{2}{\meter} threshold. This can be attributed to the fact that, in multiple cases, the positions within the hallway are mapped to the wrong floor. 
Within the Aesch building, we see a smaller decrease in performance, given that the room layout between the floors is more distinct.
\subsection{Ablation Study}
\begin{figure}[t]
  \centering
  \includegraphics[trim=0 18 0 15, clip, width=\linewidth]{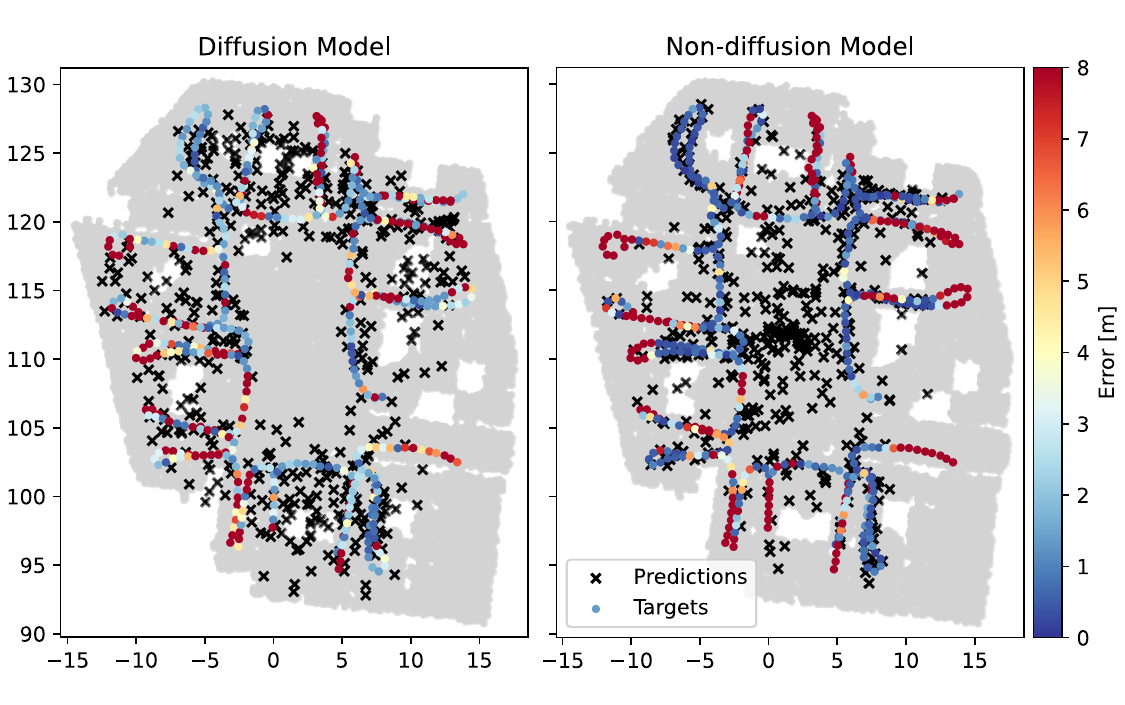}
  \caption{Predicted positions from the diffusion and non-diffusion model in Aesch Floor 3 are shown.}
  \label{fig:non_diffusion_scatter}
\end{figure}
\textbf{Regression vs Diffusion:} To understand if diffusion models are preferable to regression, we replaced the diffusion objective with direct regression on the ground truth position using the mean squared error, followed by FGR. As shown in Table~\ref{tab:ablation}, our method outperforms the non-diffusion model across most metrics and environments.
Fig.~\ref{fig:non_diffusion_scatter} illustrates the predictions of both the diffusion and regression models for the Aesch 3 floor, clearly showing the failure mode of the regression model. 
Due to the symmetrical layout, the regression model produces a large number of predictions clustered in the center of the building, given that it is not able to model the multi-modality of the scene. 

\textbf{Selection Criterion}: To identify if FGR is effective at computing a refined pose estimate, we first replace the FGR with an "Oracle" that can select the best position from the sampled $N=25$ for each sample, see Table~\ref{tab:ablation}. This removes the possible systematic error of using FGR's alignment metric to choose the best point cloud and isolates the error of the registration itself. 
Compared to multi-point prediction (Ours(N=25) + FGR), predicting a single position and applying the FGR (Ours(N=1) + FGR) performs worse. This concludes that while the FGR is worse than the Oracle at selecting the correct point cloud, it is a practical solution to select a candidate point cloud. 
\begin{figure}[t]
  \centering
  \includegraphics[trim=0 2 0 0, clip, width=0.8\linewidth]{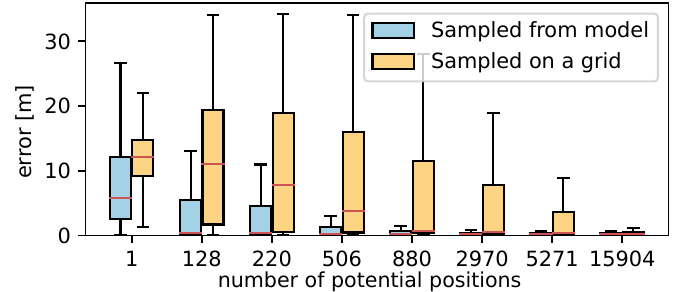}
  \caption{Error distributions for different numbers of potential positions. While sampling multiple positions from a grid, we use regular grids with resolutions of \SI{8}{\meter}, \SI{4}{\meter}, \SI{2}{\meter}, \SI{1.5}{\meter}, \SI{1.0}{\meter}, \SI{0.75}{\meter}, and  \SI{0.5}{\meter}, respectively.}
  \label{fig:boxplot}
  \vspace{-2em}
\end{figure}

\textbf{Sampling Strategy}: 
To investigate if the diffusion model is necessary for predicting the position distribution, we designed an experiment on the Aesch Floor 3 dataset, a dataset that contains multiple similar rooms. We sampled candidate positions from grids with resolutions of \SI{8}{\meter}, \SI{4}{\meter}, \SI{2}{\meter}, \SI{1.5}{\meter}, \SI{1.0}{\meter}, \SI{0.75}{\meter}, and \SI{0.5}{\meter}, while also drawing an equivalent number of samples from the diffusion model for comparison.
Fig.~\ref{fig:boxplot} illustrates the resulting error distribution for each approach. 
%
We observe that an increase in the number of samples consistently leads to a lower error rate, indicating that while there may be theoretical trade-offs, in practice, more samples prove beneficial.
Our findings show that sampling approximately 200 candidate positions from our diffusion model achieves performance comparable to sampling over 5000 points from a grid. 

The diffusion model generates samples at a rate of $1.86$ samples per second. Using FGR, the best matching candidate point cloud is selected from $25$ candidates at a speed of $5.8$ iterations per second. However, the matching time for FGR scales linearly with the number of candidates, indicating that real-time performance is only achievable using small candidate sets.

This result underscores the efficiency of our diffusion model approach, demonstrating its ability to generate high-quality candidate positions with significantly reduced computational overhead compared to exhaustive grid sampling.
\subsection{Localizability Map}
\begin{figure}[t]
  \centering
  \includegraphics[trim=0 0 0 0, clip, width=0.55\linewidth]{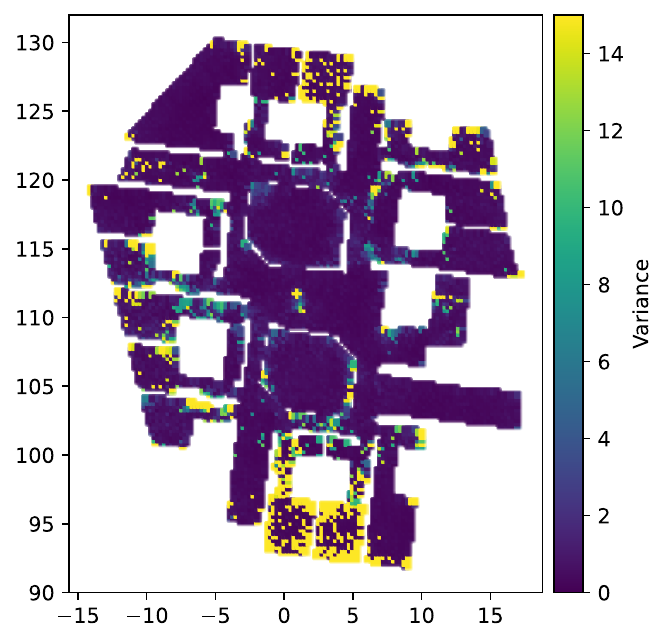}
  \caption{Localizability map of Aesch Floor 3 is provided. The color bar indicates the variance in localizability.}
  \vspace{-1em}
  \label{fig:localizationmap}
\end{figure}

In our final experiment, we explore the trained diffusion model's ability to infer LiDAR localizability within specific areas of an environment.
Intuitively, a higher variance in predictions indicates regions that are more challenging to localize accurately. To test this hypothesis, we densely sampled positions across the Aesch Floor 3 building and simulated LiDAR point clouds at each point. Using our diffusion model, we generated 8 candidate positions for each point and computed the variance of these predictions, as shown in Fig.~\ref{fig:localizationmap}.
As expected, the variance is highest in the similar apartments, indicating that the model is less certain about predictions in these areas due to their similar layouts.
Despite these initial results, the model does not perfectly detect all symmetries in the environment. We hypothesize that this is due to subtle differences in the mesh or variations in room clutter that the model, trained in simulation, is unable to exploit to identify the correct room. 
\section{Conclusion}
\label{sec:conclusion}
In this paper, we presented a novel approach to perform LiDAR-based global place recognition using a diffusion model. Our approach uses only a building mesh, which can be collected with any sensor, and the LiDAR parameters.
This allows us to successfully localize a robot using its LiDAR in challenging 
single and multi-floor construction environments. We implemented and evaluated our approach across three different datasets and showed superior performance compared to evaluated baselines.
Despite these encouraging results, this approach is sensitive to the changes in the environment that may have occured between the mesh acquisition and deployment since they are not present in the trained model.
In future work, we will further explore tackling more dynamic and changing environments by learning more robust features, e.g., by adding simulated clutter during the synthetic dataset generation.


\FloatBarrier
\bibliographystyle{plain_abbrv}
\bibliography{references}

\end{document}